\documentclass[10pt,twocolumn,letterpaper]{article}

\usepackage{cvpr}
\usepackage{times}
\usepackage{epsfig}
\usepackage{graphicx}
\usepackage{amsmath}
\usepackage{amssymb}
\usepackage{mathrsfs}
\usepackage{enumitem}
\usepackage{subfigure}
\usepackage{tabularx}
\usepackage{array}
\usepackage{multirow}
\usepackage{multicol}
\usepackage{amsmath}

\usepackage[pagebackref=true,breaklinks=true,letterpaper=true,colorlinks,bookmarks=false]{hyperref}

\cvprfinalcopy 

\def\cvprPaperID{3886} 

\ifcvprfinal\fi
\begin{document}

\title{SLADE: A Self-Training Framework For Distance Metric Learning}

\author{Jiali~Duan\textsuperscript{\rm 1}\thanks{
			\indent Work done during an internship at Amazon.
		    }\qquad~Yen-Liang~Lin\textsuperscript{\rm 2}\qquad~Son~Tran\textsuperscript{\rm 2}\qquad~Larry S.~Davis\textsuperscript{\rm 2}\qquad~C.-C. Jay Kuo\textsuperscript{\rm 1}\\
		{\textsuperscript{\rm 1} University of Southern California} ~~
		{\textsuperscript{\rm 2} Amazon}
		 \\
		\small{\texttt{\{jialidua,jckuo\}@usc.edu}} \qquad
		\small{\texttt{\{yenliang,sontran,lrrydav\}@amazon.com}} 
	}

\maketitle
\thispagestyle{empty}

\begin{abstract}
Most existing distance metric learning approaches use fully labeled data to learn the sample similarities in an embedding space. We present a self-training framework, \it{SLADE}, to improve retrieval performance by leveraging additional unlabeled data. We first train a teacher model on the labeled data and use it to generate pseudo labels for the unlabeled data. We then train a student model on both labels and pseudo labels to generate final feature embeddings. We use self-supervised representation learning to initialize the teacher model. To better deal with noisy pseudo labels generated by the teacher network, we design a new feature basis learning component for the student network, which learns basis functions of feature representations for unlabeled data. The learned basis vectors better measure the pairwise similarity and are used to select high-confident samples for training the student network. We evaluate our method on standard retrieval benchmarks: CUB-200, Cars-196 and In-shop. Experimental results demonstrate that with additional unlabeled data, our approach significantly improves the performance over the state-of-the-art methods.
\end{abstract}

\section{Introduction}
\label{sec:intro}

\begin{figure}[t]
\centering
  \includegraphics[width=0.8\linewidth]{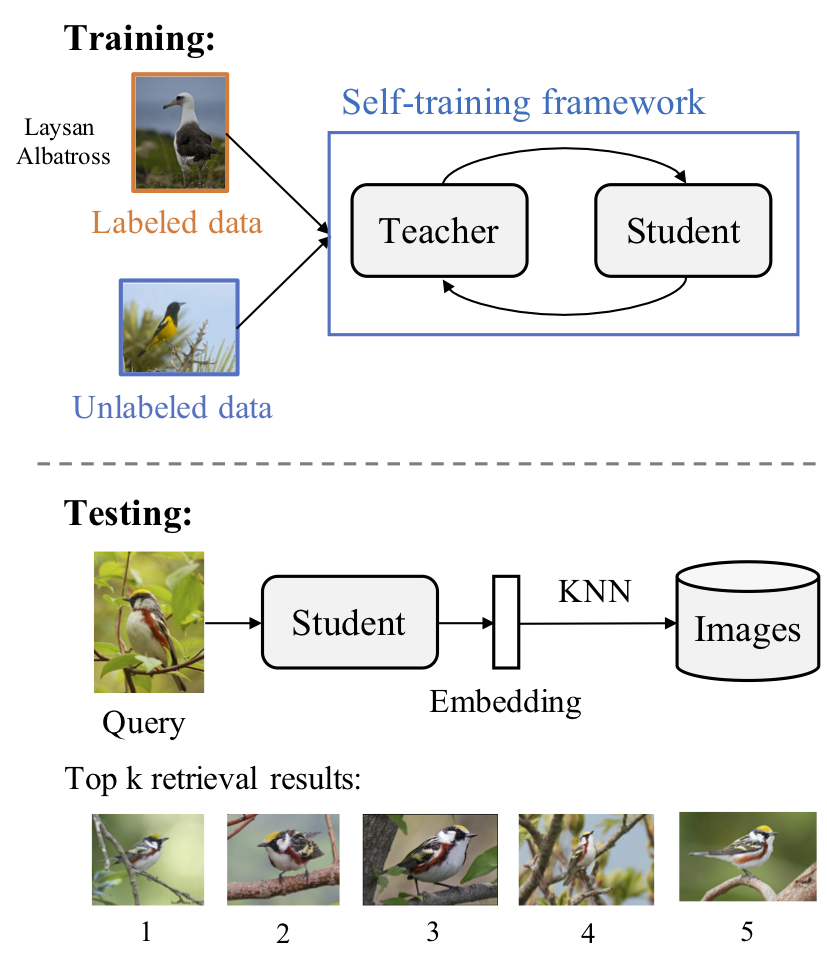}
  \caption{A self-training framework for retrieval. In the training phase, we train the teacher and student networks using both labeled and unlabeled data. In the testing phase, we use the learned student network to extract embeddings of query images for retrieval.}
  \vspace{-1.0em}
\label{fig:teaser}
\end{figure}

Existing distance metric learning methods mainly learn sample similarities and image embeddings using labeled data~\cite{oh2016deep,kim2020proxy,brown2020smooth,wang2019multi}, which often require a large amount of data to perform well. A recent study~\cite{musgrave2020metric} shows that most methods perform similarly when hyper-parameters are properly tuned despite employing various forms of losses. The performance gains likely come from the choice of network architecture. In this work, we explore another direction that uses unlabeled data to improve retrieval performance.

Recent methods in self-supervised learning ~\cite{he2020momentum,chen2020simple,caron2020unsupervised} and self-training ~\cite{xie2020self,chen2020big} have shown promising results using unlabeled data. 
Self-supervised learning leverages unlabeled data to learn general features in a task-agnostic manner. These features can be transferred to downstream tasks by fine-tuning. Recent models show that the features produced by self-supervised learning achieve comparable performance to those produced by supervised learning for downstream tasks such as detection or classification ~\cite{caron2020unsupervised}. Self-training methods~\cite{xie2020self,chen2020big} improve the performance of fully-supervised approaches by utilizing a teacher/student paradigm. However, existing methods for self-supervised learning or self-training mainly focus on classification but not retrieval.  

We present a \textbf{S}e\textbf{L}f-tr\textbf{A}ining framework for \textbf{D}istance m\textbf{E}tric learning (SLADE) by leveraging unlabeled data. Figure~\ref{fig:teaser} illustrates our method. 
We first train a teacher model on the labeled dataset and use it to generate pseudo labels for the unlabeled data.  We then train a student model on both labels and pseudo labels to generate a final feature embedding. 

We utilize self-supervised representation learning to initialize the teacher network. Most deep metric learning approaches use models pre-trained on ImageNet (\cite{kim2020proxy}, \cite{wang2019multi}, etc). Their extracted representations might over-fit to the pre-training objective such as classification and not generalize well to different downstream tasks including distance metric learning. In contrast, self-supervised representation learning ~\cite{caron2020unsupervised,chen2020simple, chen2020big, he2020momentum} learns task-neutral features and is closer to distance metric learning. For these reasons, we initialize our models using self-supervised learning approaches.
Our experimental results (Table \ref{tab:ablation_weight}) provide an empirical justification for this choice. 


Once the teacher model is pre-trained and fine-tuned, we use it to generate pseudo labels for unlabeled data. Ideally, we would directly use these pseudo labels to generate positive and negative pairs and train the student network. However in practice, these pseudo labels are noisy, which affects the performance of the student model (cf. Table~\ref{tab:ablation_comp}). Moreover, due to their different sources, it is likely that the labeled and unlabeled data include different sets of categories (see Section \ref{sec:datasets} for details about labeled and unlabeled datasets). The features extracted from the embedding layer may not adequately represent samples from those unseen classes. To tackle these issues, we propose an additional representation layer after the embedding layer. This new layer is only used for unlabeled data and aims at learning basis functions for the feature representation of unlabeled data. The learning objective is contrastive, i.e. images from the same class are mapped close while images from different classes are mapped farther apart. We use the learned basis vectors to compute the feature representation of each image and measure pairwise similarity for unlabeled data. This enables us to select high-confident samples for training the student network. Once the student network is trained, we use it to extract embeddings of query images for retrieval. 

We evaluate our model on several standard retrieval benchmarks: CUB-200, Cars-196 and In-shop. As shown in the experimental section, our approach outperforms several state-of-the-art methods on CUB-200 and Cars-196, and is competitive on In-shop. We also provide various ablation studies in the experimental section.

The main technical contributions of our work are:
\begin{itemize}
 \vspace{-1mm}\item A self-training framework for distance metric learning, which utilizes unlabeled data to improve retrieval performance. 
 \vspace{-2mm}\item A feature basis learning approach for the student network, which better deals with noisy pseudo labels generated by the teacher network on unlabeled data.
\end{itemize}


\section{Related work}
\label{sec:related}

Distance metric learning is an active research area with numerous publications. Here we review those that are relevant to our work. While a common objective is to push similar samples closer to each other and different samples away from each other, approaches differ on their losses and sample mining methods. One can train a model using cross entropy loss \cite{zhai2018classification}, hinge loss \cite{oh2016deep}, triplet loss \cite{weinberger2009distance}, proxy-NCA loss \cite{movshovitz2017no, kim2020proxy, teh2020proxynca++}, etc. \cite{movshovitz2017no} used the proxy-NCA loss to minimize the distance between a sample and their assigned anchor(s). These set of anchors were learnable. \cite{kim2020proxy} further improved the proxy-based loss by combining it with a pair-based loss. We also use a set of learnable vectors but do not optimize directly on their distances to samples. Rather, we use them as a basis (anchor set) to represent output features operated on by a distribution loss. Our intuition is that while individual pseudo labels can be noisy, representing features using these anchors makes them more robust to noise \cite{chen2020learning}, \cite{yu2019unsupervised}.

As mentioned previously, we use self-supervised training to seed our teacher model before fine-tuning it. There has been a significant progress recently in self-supervised learning of general visual representation \cite{chen2020simple, chen2020big, grill2020bootstrap, caron2020unsupervised}. \cite{chen2020simple} learns representation by minimizing the similarity between two transformed versions of the same input. Transformations include various data augmentation operations such as crop, distort or blur. \cite{chen2020big} narrowed the gap between self-supervised and supervised learning further by using larger models, a deeper projection head and a weight stabilization mechanism. In~\cite{caron2020unsupervised}, a clustering step was applied on the output presentation. Their algorithm maximizes the consistency of cluster assignments between different transformations of the same input image. Most of these works aimed at learning a generic visual representation that can later be used in various downstream tasks.


Self-training involves knowledge distillation from larger, more complex models or from ensembles of models (teachers) to less powerful, smaller students (e.g.,~\cite{hinton2015distilling, Yuan_2020_CVPR, xu2020knowledge}). Their end purpose is often reducing model size. Recently, ~\cite{xie2020self} and ~\cite{zoph2020rethinking} used iterative self-training to improve classification accuracy of both teacher and student models. At a high level, our self-training is similar to these approaches, but it is designed for distance metric learning and semi-supervised learning settings.

Unlabeled data has been used to improve performance in various computer vision tasks such as classification and semantic segmentation (e.g., \cite{ouali2020semi}). They have been also used in self-supervised and unsupervised representation learning such as in \cite{xie2020self} or \cite{zhan2020online}. But there is still a performance gap compared to the fully supervised setting. For distance metric learning, most algorithms that are competitive on popular benchmarks (CUB-200, Cars-196 and In-shop) used fully labeled data (\cite{kim2020proxy}, \cite{teh2020proxynca++}, \cite{wang2019multi}, etc). Here, we additionally use external unlabeled data to push performance on these datasets further.

\begin{figure*}[t]
\centering
  \includegraphics[width=1.0\linewidth]{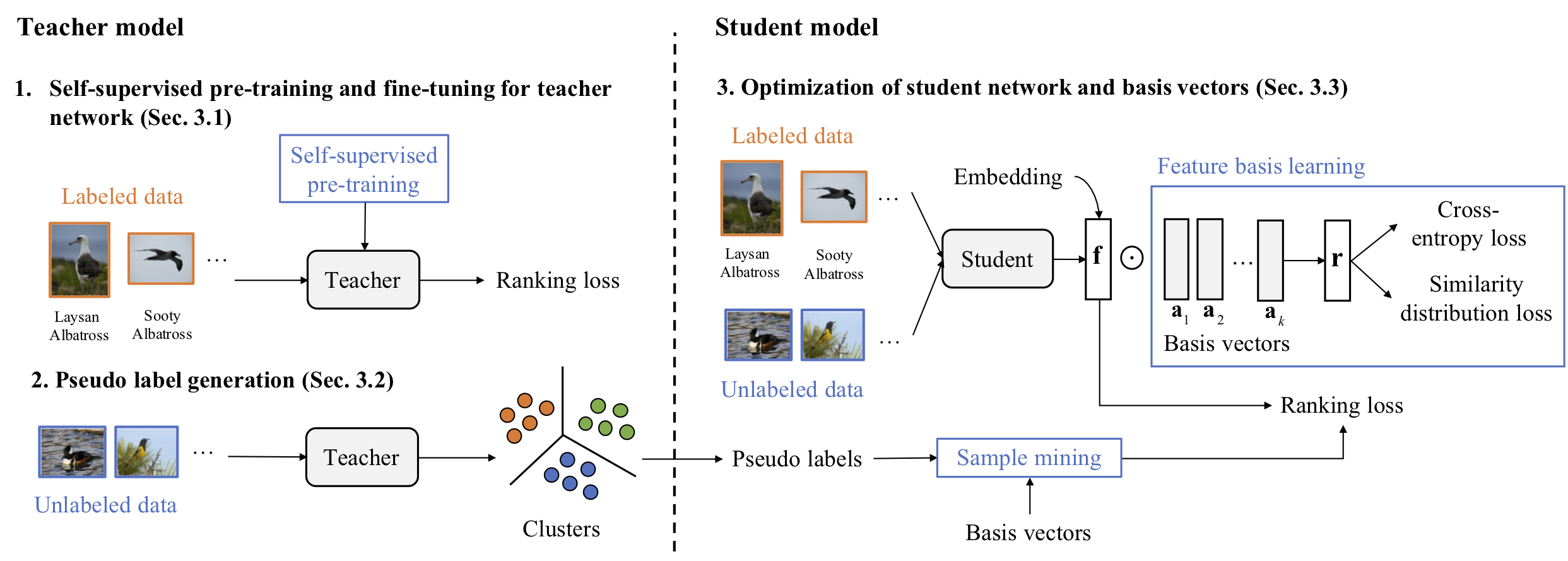}
  \caption{An overview of our self-training framework. Given labeled and unlabeled data, our framework has three main steps. (1) We first initialize the teacher network using self-supervised learning, and fine-tune it by a ranking loss on labeled data; (2) We use the learned teacher network to extract features, cluster and generate pseudo labels on unlabeled data; (3) We optimize the student network and basis vectors on labeled and unlabeled data. The purpose of feature basis learning is to select high-confidence samples (e.g., positive and negative pairs) for the ranking loss, so the student network can learn better and reduce over-fitting to noisy samples.}
\label{fig:pipeline}
\end{figure*}

\section{Method}
\label{sec:method}

Figure~\ref{fig:pipeline} illustrates the system overview of our self-training framework. Our framework has three main components. First, we use self-supervised learning to initialize the teacher network, and then fine-tune it on labeled data. We use a pre-trained model~\cite{caron2020unsupervised} and fine-tune it on our data to initialize the teacher network (we experimented with different approaches to pre-train our model. They all led to improvements over the pre-trained ImageNet model. SwAV~\cite{caron2020unsupervised} was chosen as it led to the best performance - see the experimental section). After pre-training, we fine-tune the teacher network with a ranking loss (e.g., contrastive loss) on labeled data. The details of self-supervised pre-training and fine-tuning of the teacher network are presented in section~\ref{sec:supervised_finetuning}.  

Second, we use the fine-tuned teacher network to extract features and cluster the unlabeled data using k-means clustering. We use the cluster ids as pseudo labels. In practice, these pseudo labels are noisy. Directly optimizing the student network with these pseudo labels does not improve the performance of the teacher network. Therefore, we introduce a feature basis learning approach to select high-confidence samples for training the student network. The details of pseudo label generation are presented in Section~\ref{sec:pseudo_label}.

Third, we optimize the student network and basis vectors using labeled and unlabeled data. The basis vectors are defined as a set of weights that map the feature embedding ${\bf{f}}$ of each image into a feature representation ${\bf{r}}$. We train the basis vectors such that images from the same class are mapped close and images from different classes are mapped farther apart. 
The basis vectors are used to select high-confidence samples for the ranking loss. The student network and basis vectors are optimized in an end-to-end manner. The details of student network optimization and feature basis learning are in Section~\ref{sec:optimize_student_basis}.

\subsection{Self-Supervised Pre-Training and Fine-Tuning for Teacher Network}
\label{sec:supervised_finetuning}
Existing deep metric learning methods often use the ImageNet pre-trained model \cite{deng2009imagenet} for initialization, which may over-fit to the pre-training objective, and not generalize well to downstream tasks. Instead, we use self-supervised learning to initialize the teacher model. We use self-supervised pre-trained models (~\cite{caron2020unsupervised}, \cite{chen2020mocov2}, \cite{caron2018deep}) and fine-tune them on our data. As shown in the experimental section, this choice leads to improvement in retrieval performance as compared to the pre-trained ImageNet models (see Table \ref{tab:ablation_weight}). We conjecture that this might be because deep metric learning and self-supervised learning are related, they both learn embeddings that preserve distances between similar and dissimilar data.

Specifically, we are given a set of labeled images: ${D^l} = \{ ({x_1},{y_1}),({x_2},{y_2}),...,({x_n},{y_n})\}$ and unlabeled images: ${D^u} = \{ {\hat x_1},{\hat x_2},...,{\hat x_m}\}$. We denote the parameters of the teacher network as ${\theta ^t}$ and the parameters of the student network as ${\theta ^s}$. In the pre-training stage, we fine tune the self-supervised model on the union of the labeled and unlabeled images without using the label information to initialize the teacher model. Once the teacher network is pre-trained, we fine-tune the teacher network using a ranking loss (for example, a constrastive loss \cite{hadsell2006dimensionality}) on the labeled data:

\begin{equation}
\begin{array}{c}
\begin{aligned}
{L_{rank}} = \sum\limits_{({x_i},{y_i}) \in P} {\max (d({x_i},{y_i}) - {m_{pos}},0)} \\
 + \sum\limits_{({x_i},{y_i}) \in N} \max ({m_{neg}} - d( {x_i},{y_i}),0)
\end{aligned}
\end{array}
\label{eq:rank_loss}
\end{equation}
where $P$ is the positive pairs, $N$ is the negative pairs, and ${{m_{pos}}}$ and ${{m_{neg}}}$ are the margins.

\subsection{Pseudo Label Generation}
\label{sec:pseudo_label}
We use the teacher model to extract features, and cluster the unlabeled images using k-means. The unlabeled images are assigned to the nearest cluster centers. The assigned cluster ids are then used as pseudo labels. One can train a student network with a ranking loss that uses the positive and negative pair samples sampled from the pseudo labels. 
However, in practice, the pseudo labels can be noisy, and unlabeled data may have unseen categories. The features extracted from the teacher model may not work well on those unseen categories (the pseudo labels are incorrectly estimated). To alleviate these issues, we constrain pseudo label generation to linear combinations of a set of basis vectors. These basis vectors are trained in a supervised manner using labeled data and can be considered as class centers of the labeled data~\cite{zhai2018classification}. Details are given in the next section.

\subsection{Optimization of Student Network and Basis Vectors}
\label{sec:optimize_student_basis}
We first explain the ideas of feature basis learning, and the use of basis vectors for sample mining and describe the training for student network and basis vectors. 

\subsubsection{Feature Basis Learning}
Basis vectors are a set of learnable weights that map a feature embedding ${\bf{f}}$ of an image to a feature representation ${\bf{r}}$. We denote a set of basis vectors as $\{ {{\bf{a}}_1},{{\bf{a}}_2},...,{{\bf{a}}_k}\} $, where each basis vector ${{\bf{a}}_i}$ is a $d \times 1$ vector. For simplicity, we represent the basis vectors as a $k \times d$ matrix ${{\bf{W}}_a}$. Given an image ${\bf{I}}$, we use the student network to obtain the feature embedding ${\bf{f}}$ of the input image. The feature representation ${\bf{r}}$ is computed by ${\bf{r}} = {{\bf{W}}_a} \cdot {\bf{f}}$, where $r_i^{} = {\bf{a_i}}^T \cdot {\bf{f}}$.


We train the basis vectors using two losses, a cross-entropy loss and a similarity distribution loss. The loss function for feature basis learning is defined as:
\begin{equation}
{L_{Basis}} = {L_{CE}} + {L_{SD}}
\label{eqn:anchor_learning}
\end{equation}
where the first term is the cross entropy loss on the labeled data and the second term is the similarity distribution loss on the unlabeled data.

The cross-entropy loss is applied on labeled data. The ground truth class labels can be used as a strong supervision signal to regularize the basis vectors to separate different classes. The cross entropy loss on labeled data is:
\begin{equation}
{L_{CE}} = \sum\limits_{i = 1}^n {{y_i}\log (\sigma (} {{\bf{W}}_a}f({x_i},{\theta ^s})))
\end{equation}
where $\sigma$ is the softmax function, ${{\bf{W}}_a}$ is the matrix for basis vectors, and ${\theta ^s}$ is the parameters for the student network.  
Note that since a cross-entropy loss is used here, the columns of $\bf{W}$ approximate class centers~\cite{zhai2018classification}.  Therefore, $\bf{r}$ can be viewed as a class-wise similarity representation or projections on the bases. It uses the class responses to interpolate the samples from unseen classes. This representation has also shown promising results on unlabeled data for other tasks (e.g., person Re-ID~\cite{yu2019unsupervised}).

For unlabeled data, one can also train a cross-entropy loss on the pseudo labels similar to labeled data. However, we found that this leads to poor performance since the model tends to over-fit to the noisy pseudo labels. Instead, we optimize with a global similarity distribution loss on the unlabeled data. 

%

We use the pseudo labels to sample a set of pseudo positive pairs and pseudo negative pairs, where the pseudo positive pairs are sampled from the same pseudo class and the pseudo negative pairs are sampled from different pseudo classes. We compute the similarity of each image pair by using the cosine similarity of two normalized feature representation:
\begin{equation}
    s({{\hat x}_i},{{\hat x}_j}) = \cos ({{\bf{r}}_i},{{\bf{r}}_j}) = \cos ({{\bf{W}}_a}{{\bf{f}}_i},{{\bf{W}}_a}{{\bf{f}}_j})
\label{eq:cos_sim}
\end{equation}

Directly optimizing  the  model  with  individual  similarity  based  on pseudo labels can lead to inferior performance due to noise. Instead, we  model  the  similarities  as  two  "global" Gaussian distributions $G^{+}$ and $G^{-}$ and maximize the margin between them. These distributions are less sensitive to individual pseudo label noise (results in Table \ref{tab:loss design} give an empirical justification). Specifically, we aim to separate the two Gaussian distributions by maximizing the difference between their means and penalizing the variance of each distribution. The similarity distribution loss is defined as: 
\begin{equation}
{L_{SD}}({G^ + }||{G^ - }) = \max({\mu ^ - } - {\mu ^ + } + m,0) + \lambda ({\upsilon ^ + } + {\upsilon ^ - })
\label{eqn:sdl}
\end{equation}
where $\mu^{+}$ ($\mu^{-}$) and ${\upsilon ^ + }$ (${\upsilon ^ - }$) are the mean and variance of the Gaussian distributions respectively, and $m$ is the margin. We update the parameters in a batch-wise manner:
\begin{equation}
    \begin{split}
    \mu^{+} = (1-\beta)\times \mu^{+}_b + \beta\times \mu^{+} \\
    \upsilon^{+} = (1-\beta)\times \upsilon^{+}_b + \beta\times \upsilon^{+}
    \end{split}
\label{eqn:momentum}
\end{equation}
where $\mu _b^ + $ and $\upsilon _b^ + $ are the mean and variance in a batch. $\beta$ is the updating rate. The parameters of $G^{-}$ are updated in a similar way.

\subsubsection{Sample Mining}
We use the basis vectors to select high-confidence sample pairs from the unlabeled images for training the student network. Given a set of samples in a batch, we compute the pair-wise similarity for all samples using equation \ref{eq:cos_sim} and select positive and negative pairs by:

\begin{equation}
\begin{array}{*{20}{l}}
{P = \{ ({{\hat x}_i},{{\hat x}_j})|s({{\hat x}_i},{{\hat x}_j}) \ge {T_1}\} }\\
{N = \{ ({{\hat x}_i},{{\hat x}_j})|s({{\hat x}_i},{{\hat x}_j}) \le {T_2}\} }
\label{eq:confident_pairs}
\end{array}
\end{equation}
We set the confidence thresholds of ${T_1}$ and $T_2$ using ${\mu^ + }$ and ${\mu^ - }$.
%
The positive and negative pairs will be used in the ranking function (see Equation \ref{eq:final_objective_function}).

\subsubsection{Joint Training}  
We train the student network and basis vectors by minimizing a function $L$:
\begin{equation}
\mathop {\min }\limits_{{\theta ^s},{{\bf{W}}_a}} L({\theta ^s},{{\bf{W}}_a})
\end{equation}

\begin{equation}
\begin{aligned}
L = &{L_{rank}}({D^l};{\theta ^s}) + {\lambda _1}{L_{rank}}({D^u};{\theta ^s})  \\
& +{\lambda _2}{L_{Basis}}({D^l},{D^u};{\theta ^s},{{\bf{W}}_a})
\label{eq:final_objective_function}
\end{aligned}
\end{equation}
where $L_{rank}$ is a ranking loss (see Equation \ref{eq:rank_loss}).  $\lambda_1$ and $\lambda_2$ are empirically chosen to make the magnitudes of the losses similar in scale. We train the ranking loss on both labeled and unlabeled images. Note that for unlabeled data, we use the sample mining to obtain the positive and negative pairs. 
Our framework is generic and applicable to different pair-based ranking losses. We report the results of different losses, e.g., contrastive loss and multi-similarity loss in Table \ref{tab:eval_cub_cars}.

We first train the basis vectors for a few iterations to get a good initialization, then train the student network and basis vectors end-to-end. After training the student network, we use the student as a new teacher and go back to the pseudo label generation step. We iterate this a few times. During testing, we discard the teacher model and only use the student model to extract the embedding of a query image for retrieval.

\begin{table*}[!th]
    \centering
    \begin{tabularx}
    {\textwidth} { 
    >{\centering\arraybackslash}X 
    >{\centering\arraybackslash}X 
    >{\centering\arraybackslash}X 
    >{\centering\arraybackslash}X 
    |>{\centering\arraybackslash}X >{\centering\arraybackslash}X >{\centering\arraybackslash}X
    |>{\centering\arraybackslash}X >{\centering\arraybackslash}X  >{\centering\arraybackslash}X}
    \hline
    \multicolumn{1}{l}{\multirow{2}{*}[-2mm]{Methods}} &
    \multicolumn{1}{c}{\multirow{2}{*}[-2mm]{Frwk}} & \multicolumn{1}{c}{\multirow{2}{*}[-2mm]{Init}} &
    \multicolumn{1}{c|}{\multirow{2}{*}[-2mm]{Arc / Dim}} &
    \multicolumn{3}{c|}{CUB-200-2011} & 
    \multicolumn{3}{c}{Cars-196}\\ 
    \cline{5-10}
    \multicolumn{4}{l|}{}&
        MAP@R & RP & P@1 &
        MAP@R & RP & P@1 \\ \hline 
    \multicolumn{1}{l|}{Contrastive~\cite{hadsell2006dimensionality}} &
    \multicolumn{1}{c|}{\cite{musgrave2020metric}} &
    \multicolumn{1}{c|}{ImageNet} &
    \multicolumn{1}{c|}{BN / 512} &
        26.53 & 37.24& 68.13&
        24.89 & 35.11& 81.78 \\ 
    \multicolumn{1}{l|}{Triplet~\cite{weinberger2009distance}} &
    \multicolumn{1}{c|}{\cite{musgrave2020metric}} &
    \multicolumn{1}{c|}{ImageNet} &
    \multicolumn{1}{c|}{BN / 512} &
        23.69 & 34.55& 64.24&  
        23.02 & 33.71& 79.13 \\
    \multicolumn{1}{l|}{ProxyNCA~\cite{movshovitz2017no}} &
    \multicolumn{1}{c|}{\cite{musgrave2020metric}} &
    \multicolumn{1}{c|}{ImageNet} &
    \multicolumn{1}{c|}{BN / 512} &
        24.21 & 35.14& 65.69& 
        25.38 & 35.62& 83.56 \\
    \multicolumn{1}{l|}{N. Softmax~\cite{zhai2018classification}} & 
    \multicolumn{1}{c|}{\cite{musgrave2020metric}} &
    \multicolumn{1}{c|}{ImageNet} &
    \multicolumn{1}{c|}{BN / 512}&
        25.25 & 35.99& 65.65&
        26.00 & 36.20& 83.16\\
    \multicolumn{1}{l|}{CosFace~\cite{wang2018additive,wang2018cosface}} & 
    \multicolumn{1}{c|}{\cite{musgrave2020metric}} &
    \multicolumn{1}{c|}{ImageNet} &
    \multicolumn{1}{c|}{BN / 512} &
        26.70 & 37.49& 67.32& 
        27.57 & 37.32& 85.52 \\ 
    \multicolumn{1}{l|}{FastAP~\cite{cakir2019deep}} &
    \multicolumn{1}{c|}{\cite{musgrave2020metric}} &
    \multicolumn{1}{c|}{ImageNet} &
    \multicolumn{1}{c|}{BN / 512} &
        23.53 & 34.20& 63.17& 
        23.14& 33.61 & 78.45\\
    \multicolumn{1}{l|}{MS+Miner~\cite{wang2019multi}} &
    \multicolumn{1}{c|}{\cite{musgrave2020metric}}&
    \multicolumn{1}{c|}{ImageNet}&
    \multicolumn{1}{c|}{BN / 512}&
        26.52 & 37.37 & 67.73 &
        27.01 & 37.08 & 83.67 \\ \hline
    \multicolumn{1}{l|}{ Proxy-Anchor$^{1}$~\cite{kim2020proxy}} &
    \multicolumn{1}{c|}{\cite{kim2020proxy}}&
    \multicolumn{1}{c|}{ImageNet}& 
    \multicolumn{1}{c|}{R50 / 512}&
        - & -& 69.9& 
        - & -& 87.7 \\
    \multicolumn{1}{l|}{ Proxy-Anchor$^{2}$~\cite{kim2020proxy}} &
    \multicolumn{1}{c|}{\cite{musgrave2020metric}}&
    \multicolumn{1}{c|}{ImageNet} &
    \multicolumn{1}{c|}{R50 / 512}&
        25.56 & 36.38 & 66.04 &  
        30.70 & 40.52 & 86.84 \\
    \multicolumn{1}{l|}{ ProxyNCA++~\cite{teh2020proxynca++}} &
    \multicolumn{1}{c|}{\cite{teh2020proxynca++}}&
    \multicolumn{1}{c|}{ImageNet}&
    \multicolumn{1}{c|}{R50 / 2048}&
        - & - & 72.2 & 
        - & - & 90.1 \\
    \multicolumn{1}{l|}{ Mutual-Info~\cite{ayedunifying}} &
    \multicolumn{1}{c|}{\cite{ayedunifying}} &
    \multicolumn{1}{c|}{ImageNet} &
    \multicolumn{1}{c|}{R50 / 2048} &
        - & - & 69.2 & 
        - &  -& 89.3 \\
    \hline\hline
    \multicolumn{1}{l|}{ Contrastive~\cite{hadsell2006dimensionality} (${T_1}$)} &
    \multicolumn{1}{c|}{\cite{musgrave2020metric}}&
    \multicolumn{1}{c|}{ImageNet} &
    \multicolumn{1}{c|}{R50 / 512}&
        25.02 &  35.83 & 65.28 & 
        25.97 &  36.40 & 81.22 \\
    \multicolumn{1}{l|}{ Contrastive~\cite{hadsell2006dimensionality} (${T_2}$)} & 
    \multicolumn{1}{c|}{\cite{musgrave2020metric}}&
    \multicolumn{1}{c|}{SwAV}&
    \multicolumn{1}{c|}{ R50 / 512}&
        29.29 & 39.81 & 71.15 & 
        31.73 & 41.15 & 88.07 \\
    \multicolumn{1}{l|}{ SLADE (Ours) (${S_1}$) } & 
    \multicolumn{1}{c|}{\cite{musgrave2020metric}} &
    \multicolumn{1}{c|}{ImageNet} &
    \multicolumn{1}{c|}{ R50 / 512}&
        29.38 & 40.16 & 68.92 & 
        31.38 & 40.96 & 85.8 \\
    \multicolumn{1}{l|}{ SLADE (Ours) (${S_2}$)} & 
    \multicolumn{1}{c|}{\cite{musgrave2020metric}}&
    \multicolumn{1}{c|}{SwAV} &
    \multicolumn{1}{c|}{ R50 / 512}&
         \textbf{33.59} &  \textbf{44.01} &  \textbf{73.19} & 
         \textbf{36.24} &  \textbf{44.82} &  \textbf{91.06} \\
    \hline\hline
    \multicolumn{1}{l|}{ MS ~\cite{wang2019multi} (${T_3}$)} &
    \multicolumn{1}{c|}{\cite{musgrave2020metric}}&
    \multicolumn{1}{c|}{ImageNet} &
    \multicolumn{1}{c|}{R50 / 512}&
        26.38 & 37.51 & 66.31 &  
        28.33 & 38.29 & 85.16 \\
    \multicolumn{1}{l|}{ MS ~\cite{wang2019multi} (${T_4}$)} &
    \multicolumn{1}{c|}{\cite{musgrave2020metric}}&
    \multicolumn{1}{c|}{SwAV} &
    \multicolumn{1}{c|}{R50 / 512}&
        29.22 & 40.15 & 70.81 &  
        33.42 & 42.66 & 89.33 \\
    \multicolumn{1}{l|}{ SLADE (Ours) (${S_3}$) } & 
    \multicolumn{1}{c|}{\cite{musgrave2020metric}}&
    \multicolumn{1}{c|}{ImageNet}&
    \multicolumn{1}{c|}{ R50 / 512}&
        30.90&  41.85 & 69.58& 
        32.05 & 41.50 & 87.38\\
    \multicolumn{1}{l|}{ SLADE (Ours) (${S_4}$)} & 
    \multicolumn{1}{c|}{\cite{musgrave2020metric}}&
    \multicolumn{1}{c|}{SwAV}&
    \multicolumn{1}{c|}{ R50 / 512} &
         \textbf{33.90} &  \textbf{44.36} &  \textbf{74.09}& 
         \textbf{37.98} &  \textbf{46.92} &  \textbf{91.53} \\ 
    \hline
    \end{tabularx}
    \vspace{0.1mm}
    \caption{
        MAP@R, RP, P@1 ($\%$) on the CUB-200-2011 and Cars-196 datasets. Pre-trained Image-Net model is denoted as ImageNet and the fine-tuned SwAV model on our data is denoted as SwAV. 
        The teacher networks (${T_1}$, ${T_2}$, ${T_3}$ and ${T_4}$) are trained with the different losses, which are then used to train the student networks (${S_1}$, ${S_2}$, ${S_3}$ and ${S_4}$) (e.g., the teacher ${T_1}$ is used to train the student ${S_1}$).
        Note that the results may not be directly comparable as some methods (e.g., ~\cite{kim2020proxy, teh2020proxynca++, ayedunifying}) report the results based on their own frameworks with different settings, e.g., embedding dimensions, batch sizes, data augmentation, optimizer etc. More detailed explanations are in Section \ref{sec:results}.}
    \label{tab:eval_cub_cars}
    \vspace*{-1mm}
\end{table*}

\section{Experiments}
\label{sec:exp}
We first introduce the experimental setup including datasets, evaluation criteria and implementation details. Then we report the results of our method on three common retrieval benchmarks CUB-200, Cars-196 and In-shop (\cite{wah2011caltech,krause20133d,liuLQWTcvpr16DeepFashion}) \footnote{We did not carry out experiments on the SOP dataset \cite{oh2016deep}, since we could not find an unlabeled dataset that is publicly available and is similar to it in content.}. Finally, we conduct ablation studies to analyze different design choices of the components in our framework.

\subsection{Datasets}
\label{sec:datasets}
\noindent \textbf{CUB-200/NABirds:} 
We use CUB-200-2011~\cite{wah2011caltech} as the labeled data and NABirds~\cite{van2015building} as the unlabeled data. CUB-200-2011 contains 200 fine-grained bird species, with a total number of 11,788 images. NABirds is the largest publicly available bird dataset with 400 species and 743 categories of North America's birds. It has 48,000 images with approximately 100 images for each species. We measure the overlaps between CUB-200 and NABirds, where there are 655/743 unseen classes in NABirds compared to CUB, showing the challenges of handling the out-of-domain images for unlabeled data.

\vspace{1mm}
\noindent \textbf{Cars-196/CompCars:} 
We use Cars-196~\cite{krause20133d} as the labeled data, which contains 16,185 images of 196 classes of cars. Classes are annotated at the level of make, model, and year (e.g., 2012 Tesla Model S). We use CompCars ~\cite{yang2015large} as the unlabeled data. It is collected at model level, so we filter out unbalanced categories to avoid being biased towards minority classes, resulting in 16,537 images categorized into 145 classes. 

\vspace{1mm}
\noindent \textbf{In-shop/Fashion200k:}
In-shop Clothes Retrieval Benchmark~\cite{liuLQWTcvpr16DeepFashion} includes 7,982 clothing items with 52,712 images. Different from CUB-200 and Cars196, In-shop is an instance-level retrieval task. Each article is considered as an individual category (each article has multiple views, such as front, back and side views), resulting in an average of 6.6 images per class. We use Fashion-200k~\cite{han2017automatic} as the unlabeled data, since it has similar data organization (e.g., catalog images) as the In-shop dataset.

\subsection{Evaluation Criteria} 
For CUB-200 and Cars-196, we follow the settings in ~\cite{movshovitz2017no,musgrave2020metric,kim2020proxy} that use half of the classes for training and the other half for testing, and use the evaluation protocol in~\cite{musgrave2020metric} to fairly compare different algorithms. They evaluate the retrieval performance by using MAP@R, RP and P@1. For each query, P@1 (also known as Recall@1 in previous metric learning papers) reflects whether the i-th retrieval result is correct. However, P@1 is not stable. For example, if only the first of all retrieval results is correct, P@1 is still $100\%$. RP measures the percentage of retrieval results that belong to the same class as the query. However, it does not take into account ranking of correct retrievals. $MAP@R=\frac{1}{R}\sum^{R}_{i=1}P@i$ combines the idea of mean average precision with RP and is a more accurate measure. For In-shop experiment, we use the Recall@$K$ as the evaluation metric ~\cite{liuLQWTcvpr16DeepFashion}.

We compare our model with full-supervised baselines that are trained with $100\%$ of labeled data and are fine-tuned end-to-end. Our settings are different from the ones in self-supervised learning frameworks ~\cite{chen2020simple,caron2020unsupervised}, where they evaluate the models in a label fraction setting (e.g., using $1\%$ or $10\%$ labeled data from the same dataset for supervised fine-tuning or linear evaluation). Evaluating our model in such setting is important, since in practice, we often use all available labels to fine-tune the entire model to obtain the best performance. These settings also pose several challenges. First, our fully supervised models are stronger as they are trained with $100\%$ of labels. Second, since our labeled and unlabeled data come from different image distributions, the model trained on labeled data may not work well on unlabeled data - there will be noisy pseudo labels we need to deal with. (Note that our focus is not the direct comparison to previous fully-supervised methods but to investigate the performance gain when using additional unlabeled data such as NABirds).

\subsection{Implementation Details}
We implement our model using the framework ~\cite{musgrave2020metric}. We use 4-fold cross validation and a batch size of 32 for both labeled and unlabeled data. ResNet-50 is used as our backbone network. 
%
In each fold, a student model is trained and outputs an 128-dim embedding, which will be concatenated into a 512-dim embedding for evaluation. 
We set the updating rate $\beta$ to 0.99, and $\lambda_1$ and $\lambda_2$ in Equation \ref{eq:final_objective_function} to 1 and 0.25 respectively to make the magnitude of each loss in a similar scale. 

For iterative training, we train a teacher and a student model in each fold, then use the trained student model as the new teacher model for the next fold. This produces a better teacher model more quickly compared to getting a new teacher model after the student network finishes all training folds.  

\subsection{Results}
\label{sec:results}

\begin{table}
\begin{tabularx}{0.49\textwidth} { | >{\centering\arraybackslash}X | >{\centering\arraybackslash}X >{\centering\arraybackslash}X >{\centering\arraybackslash}X >{\centering\arraybackslash}X}
    \hline
     \multicolumn{1}{l|}{Recall@$K$} & 1 & 10 & 20 & 40 \\ \hline
     \multicolumn{1}{l|}{N. Softmax~\cite{zhai2018classification}} & 88.6 & 97.5& 98.4& -\\
     \multicolumn{1}{l|}{MS~\cite{wang2019multi}} & 89.7 & 97.9& 98.5 & 99.1\\
     \multicolumn{1}{l|}{ProxyNCA++~\cite{teh2020proxynca++}} & 90.4 & 98.1& 98.8 & 99.2\\
     \multicolumn{1}{l|}{Cont. w/M~\cite{wang2020cross}} & 91.3 & 97.8& 98.4 & 99.0\\
     \multicolumn{1}{l|}{Proxy-Anchor~\cite{kim2020proxy}} & 91.5 & 98.1& 98.8 & 99.1\\
     \hline
      \multicolumn{1}{l|}{SLADE (Ours) (${S_4}$)} & 91.3& \textbf{98.6}& \textbf{99.0}& \textbf{99.4}\\
      \hline
\end{tabularx}
\vspace{0.1mm}
\caption{
    Recall@$K$ ($\%$) on the In-shop dataset. 
    }
\label{tab:eval_inshop}
\vspace{-2mm}
\end{table}

The retrieval results for CUB-200 and Cars-196 are summarized in Table~\ref{tab:eval_cub_cars}. We compare our method with state-of-the-art methods reported in~\cite{musgrave2020metric} and some recent methods~\cite{kim2020proxy,teh2020proxynca++,ayedunifying}.
Note that numbers may not be directly comparable, as some methods use their own settings. For example, ProxyAnchor~\cite{kim2020proxy} uses a larger batch size of 120 for CUB-200 and Cars-196. It also uses the combination of global average pooling and global max pooling. Mutual-Info~\cite{ayedunifying} uses a batch size of 128 and a larger embedding size of 2048. ProxyNCA++~\cite{teh2020proxynca++} uses a different global-pooling, layer normalization and data sampling scheme. 

We evaluate the retrieval performance using original images for CUB-200 and Cars-196 rather than cropped images such as in \cite{CGD} (CGD). For ProxyAnchor, in addition to reporting results using their own framework ~\cite{kim2020proxy} (denoted as ProxyAnchor$^{1}$), we also report the results using the framework~\cite{musgrave2020metric} (denoted as ProxyAnchor$^{2}$). 

We use the evaluation protocol of ~\cite{musgrave2020metric} for fair comparison. We use ResNet50 instead of BN-Inception as our backbone network because it is commonly used in self-supervised frameworks. We experiment with different ranking losses in our framework, e.g., contrastive loss ~\cite{hadsell2006dimensionality} and multi-similarity loss ~\cite{wang2019multi}. The teacher networks (${T_1}$, ${T_2}$, ${T_3}$ and ${T_4}$) are used to train the corresponding student networks (${S_1}$, ${S_2}$, ${S_3}$ and ${S_4}$) (e.g., teacher ${T_1}$ is used to train student ${S_1}$). We use the same loss for both teacher and student networks. For example, ${T_1}$ (teacher) and ${S_1}$ (student) are trained with the contrastive loss. 

We also report the results of our method using different pre-trained models: pre-trained Image-Net model (denoted as ImageNet) and the self supervised pre-trained (then fine-tuned) SwAV~\cite{caron2020unsupervised} model. Note that we experimented with different approaches to pre-train our model (S2, Table \ref{tab:eval_cub_cars}). Their performance (MAP@R) on CUB200 are  30.0\% for MoCo-v2~\cite{chen2020mocov2}, 32.19\% for Deep Cluster-v2~\cite{caron2018deep}, and 33.59\% for SwAV~\cite{caron2020unsupervised}. They all led to improvements over the pre-trained ImageNet model, which was also reported in Table \ref{tab:eval_cub_cars} (S1: 29.38\%). We chose SwAV~\cite{caron2020unsupervised} as it gives us the best performance.

We compare our method with the supervised baselines that are trained with 100\% labeled data. Even in such setting, we still observe a significant improvement using our method compared to state-of-the-art approaches that use ResNet50 or BN-Inception. We boost the final performance to $\rm{P@1} = 74.09$ and $\rm{P@1} = 91.53$ for CUB-200 and Cars-196 respectively. The results validate the effectiveness of self-supervised pre-training for retrieval as well as the feature basis learning to improve the sample quality on unlabeled data.
We also show that our method generalizes to different losses (e.g., contrastive loss ~\cite{hadsell2006dimensionality} and multi-similarity (MS) loss ~\cite{wang2019multi}). Both losses lead to improvements using our method. The performance ($\rm{P@1}$) of MS loss is improved from 66.31 to 74.09 on CUB-200, and 85.16 to 91.53 on Cars-196 respectively. We also report the performance of our method using the pre-trained ImageNet model (${S_3}$), which achieves on-par performance with state-of-the-art approaches (e.g., Proxy-Anchor) even when we use a lower baseline model as the teacher (e.g., MS loss).


Table \ref{tab:eval_inshop} summarizes the results for In-shop. Different from CUB-200 and Cars-196, In-shop is an instance-level retrieval task, where each individual article is considered as a category. Fashion200k is used as the unlabeled data. We train the teacher and student models using the multi-similarity loss~\cite{wang2019multi} similar to the settings as ${T_4}$ and ${S_4}$ in Table \ref{tab:eval_cub_cars}. We report the results using $\rm{Recall}@K$~\cite{liuLQWTcvpr16DeepFashion}. We achieve competitive results against several state-of-the-art methods. We note that the images in In-shop dataset are un-cropped while the images in Fashion200k dataset are cropped. So there exist notable distribution differences between these two datasets. We use the un-cropped version of In-shop to fairly compare with the baseline methods.


\begin{figure*}[t]
\centering
      \includegraphics[width=1.0\linewidth]{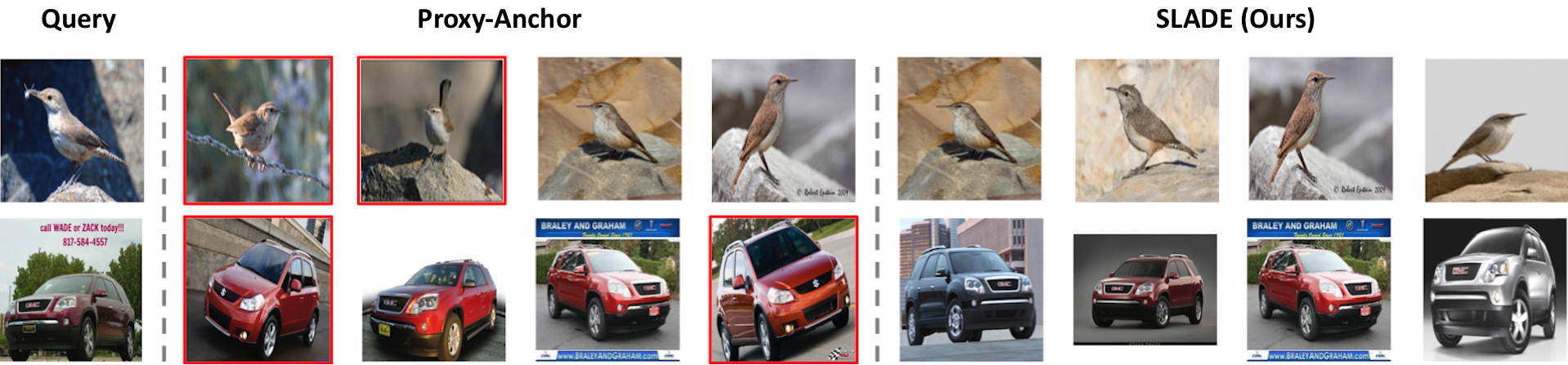}
  \caption{Retrieval results on CUB-200 and Cars-196. We show some challenging cases where our self-training method improves Proxy-Anchor \cite{kim2020proxy}. Our results are generated based on the student model ${S_2}$ in Table \ref{tab:eval_cub_cars}. The red bounding boxes are incorrect predictions.}
\label{fig:result}
\end{figure*}

\subsection{Ablation study}
%
\subsubsection{Initialization of Teacher Network}
We first investigate the impact of using different pre-trained weights to initialize the teacher network (see Table~\ref{tab:ablation_weight}). The results in the table are the final performance of our framework using different pre-trained weights. The teacher network is trained with a contrastive loss. 
We compare three different pre-trained weights: (1) a model trained on ImageNet with supervised learning; (2) a model trained on ImageNet with self-supervised approaches (specifically~\cite{caron2020unsupervised}); and (3) a self-supervised model with further fine-tuning on our data without using label information.
From the results, we can see that the weights from self-supervised learning significantly improve the pre-trained ImageNet model, $4.21\%$ and $4.86\%$ improvements on CUB-200 and Cars-196 respectively. This validates the effectiveness of self-supervised pre-training for retrieval. 
We also find that fine-tuned model (3) further improves the performance ($\sim 1\%$ improvements) of pre-trained model (2).


\label{sec:compare}

\begin{table}[h]
\centering
\begin{tabularx}{0.44\textwidth} {>{\centering\arraybackslash}X | >{\centering\arraybackslash}X |>{\centering\arraybackslash}X}
    \hline
    \multicolumn{1}{c|}{\multirow{2}{*}[0mm]{Pre-trained weight}} & \multicolumn{2}{c}{MAP@R} \\   \cline{2-3}
    & \multicolumn{1}{c|}{CUB-200} & Cars-196 \\ \hline
     \multicolumn{1}{c|} {ImageNet~\cite{deng2009imagenet}} & 29.38 & 31.38\\
     \multicolumn{1}{c|} {Pre-trained SwAV~\cite{caron2020unsupervised}} & 32.79 & 35.54 \\
     \multicolumn{1}{c|} {Fine-tuned SwAV} & 33.59&  36.24\\
    \hline
\end{tabularx}
\vspace{1mm}
\caption{Comparison of different weight initialization schemes of the teacher network, where the teacher is trained with a contrastive loss. The results are the final performance of our framework.}
\label{tab:ablation_weight}
\vspace*{-1mm}
\end{table}

\subsubsection{Components in Student Network}

Table~\ref{tab:ablation_comp} analyzes the importance of each component in our self-training framework. 
The results are based on the teacher network trained with a contrastive loss. 
Training the student with the positive and negative pairs sampled from pseudo labels only improves the teacher slightly, $1.52\%$ and $0.26\%$ on CUB-200 and Cars-196 respectively. 
The improvements are not very significant because the pseudo labels are noisy on unlabeled data.
The performance is further improved by using the feature basis learning and sample mining, which supports the proposed method for better regularizing the embedding space with the feature basis learning and selecting the high-confidence sample pairs.
We boost the final performance to 33.59 and 36.24 for CUB-200 and Cars-196 respectively. 
%

\begin{table}[h]
\centering
\begin{tabularx}{0.44\textwidth} {>{\centering\arraybackslash}X | >{\centering\arraybackslash}X |>{\centering\arraybackslash}X}
    \hline
    \multicolumn{1}{c|}{\multirow{2}{*}[0mm]{Components}} & \multicolumn{2}{c}{MAP@R} \\   \cline{2-3}
    & \multicolumn{1}{c|}{CUB-200}              & Cars-196 \\ \hline
    \multicolumn{1}{l|}{Teacher (contrastive)}  & 29.29 & 31.73\\ 
    \hline
    \multicolumn{1}{l|}{Student (pseudo label)} & 30.81 & 31.99\\
    \multicolumn{1}{l|} {+ Basis}               & 32.45&  35.78\\
    \multicolumn{1}{l|} {+ Basis + Mining}      &33.59& 36.24\\
    \hline
\end{tabularx}
\vspace{1mm}
\caption{Ablation study of different components in our framework on CUB-200 and Cars-196. The teacher network is trained with a contrastive loss.}
\label{tab:ablation_comp}
\vspace{-1mm}
\end{table}

\subsubsection{Pairwise Similarity Loss}
We also investigate different design choices of the loss functions (Equation~\ref{eqn:sdl}) used for feature basis learning. One alternative is to assign a binary label to each constructed pair according to the pseudo labels (e.g., $1$ for pseudo-positive pair and $0$ for pseudo-negative pair) and calculate a batch-wise cross entropy loss between pairwise similarities against these binary labels. We denote this option as local-CE. Another alternative is to first update the global Gaussian means using Equation~\ref{eqn:momentum}, and then assign a binary label to the global means, followed by a cross entropy loss. We denote this option as global-CE. Similarity distribution loss performs better than cross-entropy, either locally or globally. The reason could be that basis vectors need not generalize to pseudo labels as they could be noisy. 

\begin{table}[h]
\centering
\begin{tabularx}{0.44\textwidth} {>{\centering\arraybackslash}X | >{\centering\arraybackslash}X |
>{\centering\arraybackslash}X | >{\centering\arraybackslash}X}
    \hline
    \multicolumn{1}{c|}{\multirow{2}{*}[0mm]{Regularization}} & \multicolumn{3}{c}{CUB-200} \\   \cline{2-4}
    & \multicolumn{1}{c|}{MAP@R} & \multicolumn{1}{c|}{RP} & 
    \multicolumn{1}{c}{P@1} \\ \hline
    \multicolumn{1}{c|}{Local-CE}   
    & 32.69 & 43.20 & 72.64\\
    \multicolumn{1}{c|}{Global-CE}   
    & 32.23 & 42.68 & 72.45\\
    \multicolumn{1}{c|}{SD (Ours)}   
    & 33.59 & 44.01 & 73.19\\
    \hline
\end{tabularx}
\vspace{1mm}
\caption{Accuracy of our model in MAP@R, RP and P@1 versus different loss designs on CUB-200.}
\label{tab:loss design}
\vspace*{-1mm}
\end{table}

\subsubsection{Number of Clusters}

Tables \ref{tab:influence_of_k} analyzes the influence of different numbers of clusters on unlabeled data. 
The results are based on the teacher network trained with a contrastive loss.  
The best performance is obtained with $k=400$, which is not surprising, as our model (student network) is trained on the NABirds dataset, which has 400 species.  
We can also see that our method is not sensitive to the number of clusters. 


\begin{table}[h]
\centering
\begin{tabularx}{0.44\textwidth} {>{\centering\arraybackslash}X | >{\centering\arraybackslash}X |
>{\centering\arraybackslash}X | >{\centering\arraybackslash}X}
    \hline
    \multicolumn{1}{c|}{\multirow{2}{*}[0mm]{$k$}} & \multicolumn{3}{c}{NABirds} \\   \cline{2-4}
    & \multicolumn{1}{c|}{MAP@R} & \multicolumn{1}{c|}{RP} & 
    \multicolumn{1}{c}{P@1} \\ \hline
    \multicolumn{1}{c|}{100}   
    & 31.83 & 42.25& 72.19\\
    \multicolumn{1}{c|}{200}   
    & 32.61 & 43.02& 72.75\\
    \multicolumn{1}{c|}{300}   
    & 32.81 & 43.18 & 72.21\\
    \multicolumn{1}{c|}{400}   
    & 33.59 & 44.01 & 73.19\\
    \multicolumn{1}{c|}{500}   
    & 33.26 & 43.69 & 73.26\\
    \hline
\end{tabularx}
\vspace{1mm}
\caption{Influence of using different numbers of clusters ($k$) on NABirds, which is used as the unlabeled data for CUB-200.}
\label{tab:influence_of_k}
\vspace*{-1mm}
\end{table}

\section{Conclusion}
\label{sec:conclusion}
 We presented a self-training framework for deep metric learning that improves the retrieval performance by using unlabeled data. Self-supervised learning is used to initialize the teacher model. To deal with noisy pseudo labels, we introduced a new feature basis learning approach that learns basis functions to better model pairwise similarity. The learned basis vectors are used to select high-confidence sample pairs, which reduces the noise introduced by the teacher network and allows the student network to learn more effectively. Our results on standard retrieval benchmarks demonstrate our method outperforms several state-of-the art methods, and significantly boosts the performance of fully-supervised approaches.

\vspace{2mm}
\noindent \textbf{Acknowledgement:} The authors would like to thank Zhibo Yang for helpful comments on SwAV experiments. 

\clearpage
{\small
\bibliographystyle{ieee_fullname}
\bibliography{egbib}
}
\end{document}